\pdfoutput=1
\documentclass[10pt,twocolumn,letterpaper]{article}

\PassOptionsToPackage{table,dvipsnames}{xcolor}
\usepackage[pagenumbers]{cvpr}

\usepackage{amsthm}
\usepackage{array}
\usepackage{float}
\usepackage{tikz}
\usetikzlibrary{arrows.meta,positioning,shapes.geometric,calc}
\usepackage[expansion=false]{microtype}

\definecolor{cvprblue}{rgb}{0.21,0.49,0.74}
\usepackage[
  breaklinks,
  colorlinks,
  allcolors=cvprblue,
  bookmarks=true,
  bookmarksnumbered=true,
  bookmarksopen=true,
  bookmarksopenlevel=2
]{hyperref}
\usepackage{bookmark}
\bookmarksetup{depth=2,open}
\definecolor{tblhead}{HTML}{EAF0F7}
\definecolor{tblstripe}{HTML}{F7F9FC}
\definecolor{tblhi}{HTML}{E3EFFA}
\definecolor{tblgood}{HTML}{166534}
\definecolor{tblwarn}{HTML}{B42318}
\definecolor{tblmuted}{HTML}{6B7280}
\newcommand{\Res}{\mathcal{R}}
\newcommand{\Users}{\mathcal{U}}
\newcommand{\Bands}{\mathcal{T}}
\newcommand{\Beams}{\mathcal{P}}
\newcommand{\CG}{\mathcal{C}}

\newtheorem{proposition}{Proposition}
\newtheorem{lemma}{Lemma}

\floatstyle{ruled}
\newfloat{algorithm}{tbp}{loa}
\floatname{algorithm}{Algorithm}

\title{FPTR: Feasibility-Preserving Transactional Refinement for Deadline-Constrained Joint Beam and Resource Allocation}

\author{%
Minghao Kong$^{1,2}$\thanks{Minghao Kong and Zirong Liu contributed equally to this work.}\quad
Zirong Liu$^{1\star}$\quad
Jianhao Wu$^{1}$\quad
Muyao Chen$^{1}$\quad
Jianxin Wu$^{2}$\thanks{Corresponding authors: Jianxin Wu and Rongjie Wang.}\quad
Rongjie Wang$^{1\dagger}$\\[6pt]
{\normalsize $^{1}$Department of Network Intelligence, Pengcheng Laboratory, Shenzhen, China}\\
{\normalsize $^{2}$School of Electronics and Communication Engineering, Sun Yat-sen University, Shenzhen Campus, Shenzhen, China}\\[2pt]
{\tt\small wangrj@pcl.ac.cn \qquad wujx65@mail.sysu.edu.cn}
}

\hypersetup{
  pdftitle={FPTR: Feasibility-Preserving Transactional Refinement for Deadline-Constrained Joint Beam and Resource Allocation},
  pdfauthor={Minghao Kong, Zirong Liu, Jianhao Wu, Muyao Chen, Jianxin Wu, Rongjie Wang}
}

\begin{document}

% Title-footnote symbols: shared first authors (*), corresponding authors (dagger).
\begingroup
\makeatletter
\renewcommand{\@fnsymbol}[1]{\ensuremath{\ifcase#1\or\star\or\dagger\else\@ctrerr\fi}}
\makeatother
\maketitle
\endgroup

\begin{abstract}
Short-interval wireless schedulers must deliver a complete, feasible beam-and-resource allocation before each execution deadline; any late or invalid output yields zero utility in the current control cycle.
Deeper search can raise transmitted traffic, but it also risks overrunning the deadline or leaving an infeasible intermediate state.
To resolve this tension, we propose \emph{Feasibility-Preserving Transactional Refinement} (FPTR), a framework that retains a validated feasible allocation---the \emph{incumbent}---throughout execution.
Starting from an empty fallback and a feasible base constructor, each refinement stage builds an isolated candidate without modifying the incumbent.
A candidate replaces the incumbent only if it is complete, passes feasibility validation before its stage cutoff, and strictly improves the objective; otherwise, it is discarded and the incumbent remains intact.
Four bounded refinement stages operate under this shared gate: buffer-aware resource selection, compatibility-group (CG) user sharing, demand-driven beam-mask rebuilding, and joint two-resource reassignment.
Across 150 fixed synthetic instances, the full pipeline (Full) raises mean instance-relative traffic by 3.83\% over a lightweight beam-first reference and by 1.79\% over its Base constructor, while returning feasible allocations in all 750 runs with zero misses of the 100-ms deadline.
In a separate suite with matched inputs where late outputs receive zero utility, Full outperforms six independent search baselines---ALNS, tabu search, GA, simulated annealing, iterated local search, and GRASP---by 6.68--8.35\%.
By decoupling mandatory feasible construction from optional refinement, FPTR converts spare computation time into verified gains without endangering the incumbent; open-source code and sealed data support reproduction.
\end{abstract}

\section{Introduction}
\label{sec:intro}

In each scheduling interval, the base-station scheduler studied here must decide which beams to activate and which users to assign to each resource within a 100-ms execution deadline.
The returned allocation must be complete, satisfy every coupling constraint, and arrive before the next control decision takes effect.
Any output that misses the deadline or violates even a single constraint contributes no traffic to that scheduling interval.

Exploring additional beam masks, user groups, or local repairs can substantially improve transmitted traffic.
Every search step, however, draws from the same execution window required for validation and serialization.
If in-place search runs past its budget or halts in an infeasible intermediate state, the scheduler risks losing a valid allocation altogether.
The central algorithmic challenge is therefore to exploit spare computation time for quality gains without ever putting a releasable feasible allocation at risk.

\begin{figure*}[t]
\centering
\includegraphics[
  width=0.90\textwidth,
  trim=100 80 100 165,
  clip
]{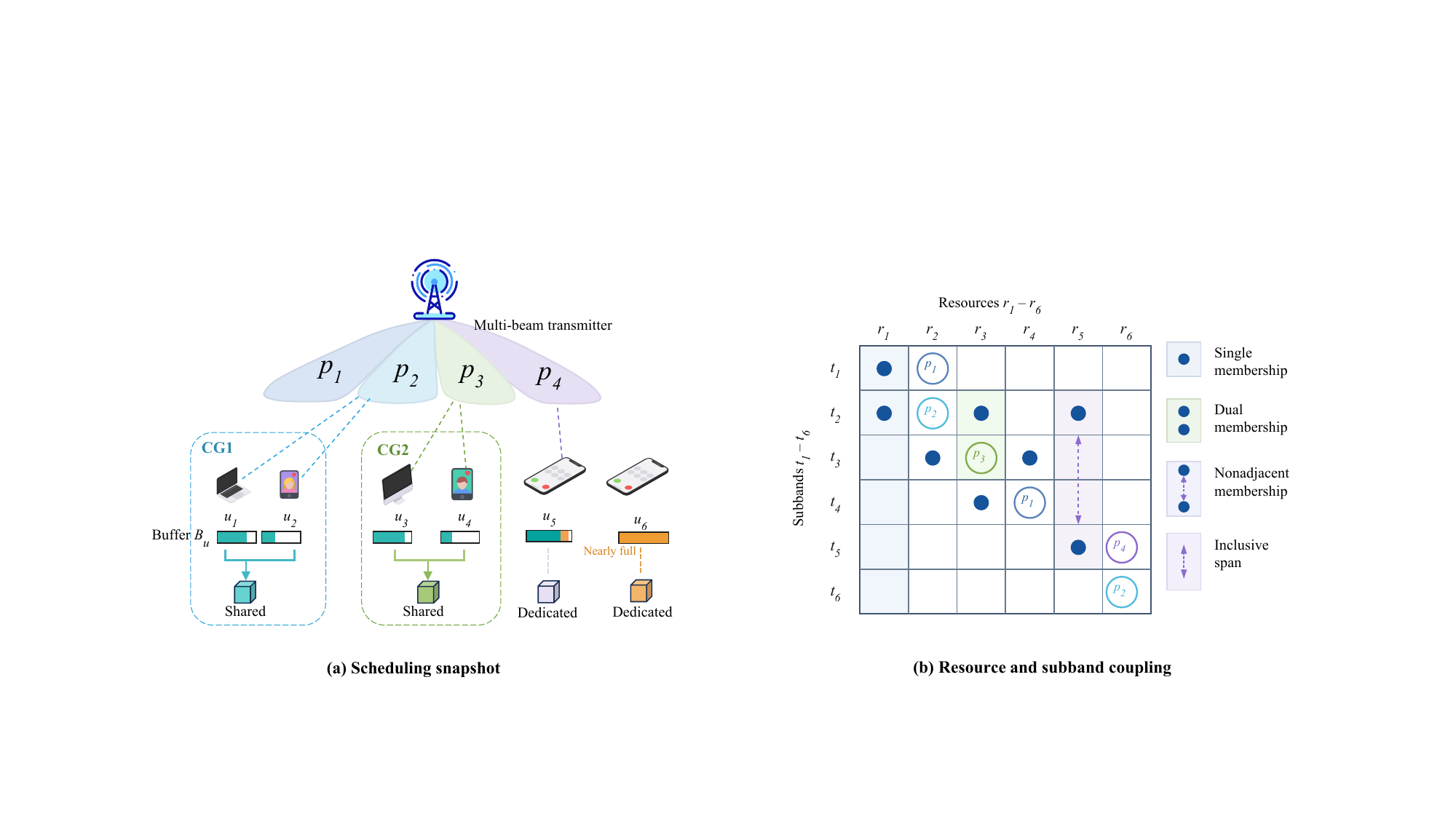}
\caption{Schematic overview of one scheduling interval. Beam activation, user assignment, finite demand, and compatibility-group sharing must be resolved jointly before the execution deadline.}
\label{fig:scenario}
\end{figure*}

Three structural couplings make this problem challenging (Fig.~\ref{fig:scenario}).
First, beam activations across all subbands and resources draw from a single global budget.
Second, finite user buffers couple assignment decisions across resources: serving a user on one resource reduces its residual demand and changes the marginal value of subsequent resources.
Third, compatibility-group (CG) sharing restricts multi-user multiplexing on a resource to users within the same group while imposing a cardinality-dependent rate penalty.
Together, binary beam masks and set-valued user assignments create a combinatorial search space in which single-pass heuristics miss profitable combinations, yet exhaustive search is prohibitive within a 100-ms window.

FPTR addresses this trade-off through one guiding principle: \emph{never put the incumbent at risk}.
The framework keeps a complete, validated allocation ready for immediate output while each optional stage constructs an isolated candidate without altering the incumbent.
A candidate replaces the incumbent only after it is complete, validated before its stage cutoff, and strictly better in objective value; if any check fails, the candidate is discarded and the incumbent survives unchanged.
This commit-or-discard gate cleanly separates mandatory feasible construction from optional quality refinement.
Throughout this paper, we use ``transactional'' strictly in this single-threaded, all-or-nothing sense: every candidate is either committed as a whole or discarded as a whole.

We make four contributions:
\begin{itemize}
  \item We formulate deadline-constrained joint beam and resource allocation as a finite combinatorial problem whose solver must preserve a valid output while optional refinement proceeds.
  \item We design FPTR, a unified commit-or-discard architecture combining a feasible base constructor with four bounded refinement stages.
  \item We prove that for a single resource under fixed beam masks, the optimal CG subset of any fixed cardinality is obtained exactly by choosing the users with the largest buffer-aware marginal gains, and we establish the feasible-incumbent invariant.
  \item We evaluate transmitted traffic, feasibility, stage acceptance, quality--time trade-offs, CG stress, matched external-baseline comparisons, and exact small-case gaps on fixed synthetic suites with published checksums.
\end{itemize}

\section{Related Work}
\label{sec:related}

Joint beam and resource allocation has been studied across terrestrial, satellite, and integrated wireless architectures.
Classical multiuser OFDM jointly allocates subcarriers, bits, and power to minimize transmit power~\cite{wong1999}; in integrated terrestrial--satellite networks, hybrid beamforming and user scheduling are co-designed for sum rate and energy efficiency~\cite{peng2021}, while deep reinforcement learning addresses fair massive-MIMO scheduling~\cite{an2023}.

In multi-beam low-Earth-orbit (LEO) systems, beam decisions couple tightly with spectrum, time, and power control.
Recent work applies online convex optimization to coordinate beam hopping, user association, subcarriers, and power~\cite{liang2025}, or optimizes adaptive beam patterns and timeslots to minimize power under demand constraints~\cite{kim2025}.
Related studies balance multi-satellite load and avoid interference~\cite{zhaobhra2025}, coordinate inter-satellite frequency reuse~\cite{wang2024}, and dynamically allocate time--frequency--space resources to close supply--demand gaps~\cite{tian2024}.
For uneven traffic and large arrays, spatial--temporal beam-pattern selection~\cite{lei2024} and effective-interference approximations~\cite{zhao2026} maintain tractability.

User grouping and sharing add further combinatorial coupling, pairing user grouping with beam scheduling~\cite{xugroup2025}, graph clustering with two-stage beamforming~\cite{song2024}, distributed fair scheduling with precoding~\cite{changfair2025}, NOMA--ODDM grouping with power and beamforming~\cite{sultana2025}, or interference-aware weighted graph partitioning~\cite{lu2026}.
Service-, energy-, and learning-driven extensions employ two-step genetic and greedy beam hopping~\cite{taksande2026}, image-assisted deep reinforcement learning under unknown demand~\cite{xuservice2026}, hierarchical reinforcement learning for age of information~\cite{xie2025}, GNN--Lyapunov cell-free mmWave scheduling~\cite{changedge2026}, and fast data- and model-driven successive convex approximation for multi-cell MIMO-OFDM~\cite{zhaoupsca2026}.
Although these approaches broaden service objectives and reduce average runtime, they focus on final schedule quality and solver cost rather than defining a release-safe incumbent when iterative refinement is interrupted or rejected.

Deadlines in wireless scheduling likewise primarily constrain packet delivery---via time-ordered LEO queues~\cite{zheng2025}, deadline-aware virtual queues~\cite{xugroup2025}, URLLC link adaptation~\cite{pocovi2018}, and throughput-optimal queue stability~\cite{tassiulas1992}---rather than when the solver itself must emit a feasible allocation.
Large-neighborhood search provides rich ruin-and-recreate moves~\cite{pisinger2019}, yet a candidate generator alone does not govern which intermediate state is safe to release upon timeout.
FPTR fills this execution-level gap: it wraps any bounded heuristic or learned generator operating on an isolated state, using an independent validator and shared commit-or-discard gate to protect the incumbent rather than universally replacing domain-specific search methods.

\section{System Model and Problem Formulation}
\label{sec:model}

Let \(\Users,\Res,\Bands,\Beams\) denote the sets of users, resources, subbands (indexed \(1,\ldots,T\)), and beams, with cardinalities \(N,K,T,P\), respectively.
Each user \(u\in\Users\) has a finite traffic buffer \(1\le B_u\le10{,}000\) (measured in snapshot traffic units), a base link quality \(\gamma_u\) (in dB), and a positive beam capability \(c_{u,p}>0\) on each beam \(p\in\Beams\).
Each resource \(r\in\Res\) belongs to an explicit subband membership set \(\Bands_r\subseteq\Bands\) with \(|\Bands_r|\in\{1,2\}\); dual-subband memberships need not be adjacent.
Its inclusive subband span is
\begin{equation}
  d_r=\max\Bands_r-\min\Bands_r+1.
  \label{eq:span}
\end{equation}
In each snapshot, the scheduler chooses binary beam-activation indicators \(\alpha_{t,p}\in\{0,1\}\) for every subband--beam pair \((t,p)\) and assigns a user subset \(G_r\subseteq\Users\) to each resource \(r\).
These decisions must satisfy the global beam budget \(L\) and the resource-activation constraints:
\begin{equation}
\begin{aligned}
  &\sum_{t\in\Bands}\sum_{p\in\Beams}\alpha_{t,p}\le L,\\
  &G_r\ne\varnothing\Rightarrow
  \sum_{p\in\Beams}\alpha_{t,p}\ge 1,\quad \forall t\in\Bands_r.
\end{aligned}
\label{eq:beam}
\end{equation}

Let \(\CG_1,\ldots,\CG_M\subseteq\Users\) be mutually disjoint compatibility groups (CGs).
Any user may occupy a resource alone, whereas two or more users may share a resource only if they all belong to the same CG:
\begin{equation}
\begin{aligned}
  G_r\in{}&\{\varnothing\}\cup\{\{u\}:u\in\Users\}\\
  &\cup\bigcup_{m=1}^{M}\{G:G\subseteq\CG_m,\ |G|\ge 2\}.
\end{aligned}
\label{eq:legal-group}
\end{equation}
A user may appear at most once in any \(G_r\), but may be scheduled across multiple resources.
For each scheduled user \(u\in G_r\), constraint~\eqref{eq:beam} and \(c_{u,p}>0\) ensure that its selected capability \(\bar c_{u,r}(\alpha)\) is strictly positive, yielding the effective link value \(z_{u,r}\) (in dB):
\begin{align}
  \bar c_{u,r}(\alpha)
  &=\frac{\displaystyle\sum_{t\in\Bands_r}\sum_{p\in\Beams}
  c_{u,p}\alpha_{t,p}}{d_r}, \label{eq:selected-cap}\\
\begin{split}
  z_{u,r}
  &=\gamma_u+10\log_{10}\frac{1}{|G_r|}\\
  &\quad+10\log_{10}\frac{\bar c_{u,r}(\alpha)}
  {\sum_{p\in\Beams}c_{u,p}}.
\end{split} \label{eq:effective-link}
\end{align}
Only the subbands explicitly listed in \(\Bands_r\) contribute to the numerator of~\eqref{eq:selected-cap}, whereas the denominator \(d_r\) spans all indices from \(\min\Bands_r\) to \(\max\Bands_r\) (for example, \(\Bands_r=\{2,5\}\) gives \(d_r=4\)).
Table~\ref{tab:traffic-map} specifies the stepwise mapping from effective link value \(z\) to single-snapshot traffic \(q(z)\).

\begin{table}[t]
\centering
\scriptsize
\renewcommand{\arraystretch}{1.15}
\setlength{\tabcolsep}{2.4pt}
\resizebox{\columnwidth}{!}{%
\begin{tabular}{lccccccc}
\toprule
\rowcolor{tblhead}
\textbf{\(z\) (dB)} & \(({-}\infty,{-}10]\) & \(({-}10,0]\) & \((0,3]\) & \((3,10]\) & \((10,15]\) & \((15,20]\) & \((20,\infty)\)\\
\midrule
\textbf{\(q(z)\)} \(\uparrow\) & \textcolor{tblmuted}{\textit{0}} & 8 & 24 & 90 & 120 & \underline{162} & \cellcolor{tblhi}\textbf{222}\\
\bottomrule
\end{tabular}%
}
\caption{Stepwise mapping from effective link value \(z\) (dB) to one-snapshot traffic \(q(z)\). \textcolor{tblmuted}{\textit{Outage}} (\textit{0}) and \textbf{peak} (\textbf{222}) tiers are highlighted.}
\label{tab:traffic-map}
\end{table}

This step function provides a deterministic one-snapshot traffic abstraction rather than a continuous physical-layer rate law.
Because each user can transmit no more than its buffered demand \(B_u\), the total clipped transmitted traffic is
\begin{equation}
  F(\alpha,G)=\sum_{u\in\Users}
  \min\left\{B_u,\sum_{r:u\in G_r}q(z_{u,r})\right\}.
  \label{eq:objective}
\end{equation}

The resulting allocation problem is the finite combinatorial program
\begin{equation}
\begin{aligned}
  \max_{\alpha,G}\quad &F(\alpha,G)\\
  \text{s.t.}\quad
  &\eqref{eq:beam},\ \eqref{eq:legal-group},\qquad
  \alpha_{t,p}\in\{0,1\}\ \ \forall(t,p).
\end{aligned}
\label{eq:problem}
\end{equation}
Operationally, a solution earns utility only if it is constraint-feasible and serialized before the external execution deadline \(D\).
Even before enforcing the beam budget, the \(PT\) binary activations span \(2^{PT}\) candidate masks, of which the budget leaves \(\sum_{\ell=0}^{L}\binom{PT}{\ell}\).
Conditioned on a mask, each enabled resource must select the empty set, a singleton, or a legal subset of one CG.
Because buffer clipping and cardinality-dependent sharing shift user utilities dynamically across resources, static user rankings are unreliable.
We therefore use bounded refinement on full-size instances and exact optimization on small cases.

\section{Feasibility-Preserving Transactional Refinement}
\label{sec:fptr}

\begin{figure*}[t]
\centering
\begingroup
\renewcommand{\rmdefault}{ptm}\rmfamily
% Scheme B: Restrained academic palette harmonized with Figs. 3--4 and manuscript tables
\definecolor{f2canvasbg}{HTML}{FAFBFC}
\definecolor{f2canvasbd}{HTML}{CBD5E1}
\definecolor{f2ink}{HTML}{1E293B}
\definecolor{f2muted}{HTML}{526071}
\definecolor{f2nodebd}{HTML}{64748B}
\definecolor{f2cardbd}{HTML}{CBD5E1}
\definecolor{f2gridmute}{HTML}{E2E8F0}
\definecolor{f2tailbg}{HTML}{F3F6FA}
\definecolor{f2tailbd}{HTML}{94A3B8}
\definecolor{f2badgebg}{HTML}{F1F5F9}

% Panel (a) stage accents: exact Morandi tones from plot_paper_results.py (Figs. 3--4)
\definecolor{f2s1bd}{HTML}{494A73}   % Base   (slate-indigo)
\definecolor{f2s1pill}{HTML}{EFEFF6}
\definecolor{f2s2bd}{HTML}{6F7FA6}   % Global (steel-blue)
\definecolor{f2s2pill}{HTML}{F0F4F9}
\definecolor{f2s3bd}{HTML}{3F817F}   % CG     (slate-teal)
\definecolor{f2s3pill}{HTML}{EDF6F5}
\definecolor{f2s4bd}{HTML}{9A6B80}   % Remask (dusty-mauve)
\definecolor{f2s4pill}{HTML}{F7F0F3}
\definecolor{f2s5bd}{HTML}{A63D38}   % Pair   (brick-crimson / Full)
\definecolor{f2s5pill}{HTML}{FBF0EF}

% Panel (b) decoupled semantic focal accents matching manuscript tables
\definecolor{f2gatebd}{HTML}{2B5B84} % Deep steel-blue focal anchor
\definecolor{f2gatebg}{HTML}{EAF0F7} % Matches tblhead
\definecolor{f2gatehl}{HTML}{DCE6F2}

\definecolor{f2passbd}{HTML}{2F6F4E} % Muted forest green
\definecolor{f2passbg}{HTML}{EDF7F1}
\definecolor{f2passtx}{HTML}{166534} % Matches tblgood

\definecolor{f2failbd}{HTML}{B42318} % Matches tblwarn
\definecolor{f2failbg}{HTML}{FDF2F2}
\definecolor{f2failtx}{HTML}{991B1B}

\tikzset{
  pics/f2watch/.style={
    code={
      \draw[line width=0.8pt,color=#1] (0,0) circle (0.14);
      \draw[line width=0.7pt,color=#1] (0,0) -- (0,0.085);
      \draw[line width=0.7pt,color=#1] (0,0) -- (0.065,-0.025);
      \draw[line width=0.9pt,color=#1] (-0.045,0.175) -- (0.045,0.175);
      \draw[line width=0.8pt,color=#1] (0,0.14) -- (0,0.175);
      \draw[line width=0.7pt,color=#1] (0.095,0.105) -- (0.135,0.14);
    }
  },
  pics/f2doc/.style={
    code={
      \draw[line width=0.7pt,color=#1,fill=white,rounded corners=0.6pt]
        (-0.11,-0.14) -- (-0.11,0.14) -- (0.04,0.14) -- (0.11,0.07) -- (0.11,-0.14) -- cycle;
      \draw[line width=0.6pt,color=#1] (0.04,0.14) -- (0.04,0.07) -- (0.11,0.07);
      \draw[line width=0.55pt,color=#1] (-0.065,0.02) -- (0.065,0.02);
      \draw[line width=0.55pt,color=#1] (-0.065,-0.03) -- (0.065,-0.03);
      \draw[line width=0.55pt,color=#1] (-0.065,-0.08) -- (0.03,-0.08);
    }
  },
  pics/f2user/.style={
    code={
      \draw[line width=0.75pt,color=#1,fill=white] (0,0.075) circle (0.055);
      \draw[line width=0.75pt,color=#1,fill=white] (-0.10,-0.09) arc[start angle=180,end angle=0,x radius=0.10,y radius=0.095] -- cycle;
    }
  },
  f2arrow/.style={-{Stealth[length=5.2pt,width=4pt]}, line width=0.9pt, color=f2ink},
  f2stagearrow/.style={-{Stealth[length=5pt,width=3.8pt]}, line width=0.85pt, color=f2nodebd},
}
\resizebox{\textwidth}{!}{%
\begin{tikzpicture}[font=\rmfamily, >=Stealth]

% =========================================================================
% PANEL (a): Ordered stages and cumulative cutoffs (width = 17.4 cm)
% =========================================================================
\draw[rounded corners=6pt, dashed, line width=0.8pt, color=f2canvasbd]
  (0, 0) rectangle (17.4, 4.42);

\node[anchor=west, font=\rmfamily\bfseries\small, text=f2ink]
  at (0.32, 4.08) {(a) Ordered stages and cumulative cutoffs};

% 5 cards: width = 3.04, gap = 0.39 -> 5*3.04 + 4*0.39 = 16.76, margin = 0.32
% Card x-origins: 0.32, 3.75, 7.18, 10.61, 14.04
\def\cw{3.04}
\def\ch{2.80}
\def\cy{0.94}

% --- Stage 1: Base ---
\begin{scope}[shift={(0.32, \cy)}]
  \begin{scope}
    \clip[rounded corners=5pt] (0,0) rectangle (\cw,\ch);
    \fill[white] (0,0) rectangle (\cw,\ch);
    \fill[f2s1bd] (0, \ch-0.11) rectangle (\cw, \ch);
  \end{scope}
  \draw[rounded corners=5pt, line width=0.75pt, color=f2cardbd] (0,0) rectangle (\cw,\ch);
  \node[font=\rmfamily\bfseries\small, text=f2ink] at (\cw/2, 2.43) {Base};
  \pic at (0.94, 2.00) {f2watch=f2s1bd};
  \node[rounded corners=3pt, draw=f2s1bd!65, fill=f2s1pill, line width=0.55pt,
        font=\rmfamily\bfseries\scriptsize, text=f2s1bd, inner xsep=5.5pt, inner ysep=2.2pt]
        at (1.78, 2.00) {45 ms};
  % Icon: 4x4 beam-resource grid + stage-accented checkmark badge
  \begin{scope}[shift={(1.12, 0.86)}]
    \fill[white] (0,0) rectangle (0.80,0.80);
    \fill[f2s1bd!22] (0,0.40) rectangle (0.20,0.60);
    \fill[f2s1bd!22] (0.40,0.60) rectangle (0.60,0.80);
    \fill[f2s1bd!22] (0.20,0.20) rectangle (0.40,0.40);
    \fill[f2s1bd!22] (0.60,0.40) rectangle (0.80,0.60);
    \draw[line width=0.55pt, color=f2nodebd] (0,0) grid[step=0.20] (0.80,0.80);
    \fill[f2s1bd] (0.78,0.06) circle (0.14);
    \draw[white, line width=1.05pt, line cap=round, line join=round]
      (0.715,0.06) -- (0.76,0.015) -- (0.85,0.11);
  \end{scope}
  \node[align=center, font=\rmfamily\scriptsize, text=f2muted, inner sep=1pt]
    at (\cw/2, 0.36) {Diversified-mask\\feasible construction};
\end{scope}

% --- Stage 2: Global ---
\begin{scope}[shift={(3.75, \cy)}]
  \begin{scope}
    \clip[rounded corners=5pt] (0,0) rectangle (\cw,\ch);
    \fill[white] (0,0) rectangle (\cw,\ch);
    \fill[f2s2bd] (0, \ch-0.11) rectangle (\cw, \ch);
  \end{scope}
  \draw[rounded corners=5pt, line width=0.75pt, color=f2cardbd] (0,0) rectangle (\cw,\ch);
  \node[font=\rmfamily\bfseries\small, text=f2ink] at (\cw/2, 2.43) {Global};
  \pic at (0.94, 2.00) {f2watch=f2s2bd};
  \node[rounded corners=3pt, draw=f2s2bd!65, fill=f2s2pill, line width=0.55pt,
        font=\rmfamily\bfseries\scriptsize, text=f2s2bd, inner xsep=5.5pt, inner ysep=2.2pt]
        at (1.78, 2.00) {60 ms};
  % Icon: neutral bars + single stage-highlighted tallest bar & selection arc
  \begin{scope}[shift={(1.01, 0.84)}]
    \draw[line width=0.6pt, color=f2nodebd, fill=f2gridmute, rounded corners=0.8pt] (0.00,0) rectangle (0.18,0.40);
    \draw[line width=0.6pt, color=f2nodebd, fill=f2gridmute, rounded corners=0.8pt] (0.27,0) rectangle (0.45,0.52);
    \draw[line width=0.75pt, color=f2s2bd, fill=f2s2bd!22, rounded corners=0.8pt] (0.54,0) rectangle (0.72,0.68);
    \draw[line width=0.6pt, color=f2nodebd, fill=f2gridmute, rounded corners=0.8pt] (0.81,0) rectangle (0.99,0.34);
    \draw[->, line width=0.85pt, color=f2s2bd] (0.10,0.48) .. controls (0.24,0.84) and (0.54,0.86) .. (0.63,0.72);
    \draw[line width=0.6pt, color=f2nodebd] (-0.06,0) -- (1.06,0);
  \end{scope}
  \node[align=center, font=\rmfamily\scriptsize, text=f2muted, inner sep=1pt]
    at (\cw/2, 0.36) {Buffer-aware\\marginal selection};
\end{scope}

% --- Stage 3: CG ---
\begin{scope}[shift={(7.18, \cy)}]
  \begin{scope}
    \clip[rounded corners=5pt] (0,0) rectangle (\cw,\ch);
    \fill[white] (0,0) rectangle (\cw,\ch);
    \fill[f2s3bd] (0, \ch-0.11) rectangle (\cw, \ch);
  \end{scope}
  \draw[rounded corners=5pt, line width=0.75pt, color=f2cardbd] (0,0) rectangle (\cw,\ch);
  \node[font=\rmfamily\bfseries\small, text=f2ink] at (\cw/2, 2.43) {CG};
  \pic at (0.94, 2.00) {f2watch=f2s3bd};
  \node[rounded corners=3pt, draw=f2s3bd!65, fill=f2s3pill, line width=0.55pt,
        font=\rmfamily\bfseries\scriptsize, text=f2s3bd, inner xsep=5.5pt, inner ysep=2.2pt]
        at (1.78, 2.00) {70 ms};
  % Icon: neutral dashed Same CG box with 3 users -> stage-accented Shared pill
  \begin{scope}[shift={(0.62, 0.76)}]
    \draw[rounded corners=3pt, dashed, line width=0.65pt, color=f2s3bd!65, fill=f2canvasbg]
      (0,0) rectangle (1.80,1.02);
    \node[font=\rmfamily\bfseries\tiny, text=f2s3bd] at (0.90, 0.89) {Same CG};
    \pic at (0.36, 0.56) {f2user=f2nodebd};
    \pic at (0.90, 0.56) {f2user=f2nodebd};
    \pic at (1.44, 0.56) {f2user=f2nodebd};
    \draw[line width=0.55pt, color=f2nodebd] (0.36,0.44) -- (0.36,0.34) -- (1.44,0.34) -- (1.44,0.44);
    \draw[line width=0.55pt, color=f2nodebd] (0.90,0.44) -- (0.90,0.26);
    \draw[rounded corners=1.5pt, line width=0.65pt, color=f2s3bd, fill=f2s3pill]
      (0.56,0.08) rectangle (1.24,0.26);
    \node[font=\rmfamily\tiny\bfseries, text=f2s3bd] at (0.90,0.17) {Shared};
  \end{scope}
  \node[align=center, font=\rmfamily\scriptsize, text=f2muted, inner sep=1pt]
    at (\cw/2, 0.36) {CG-focused masks\\and valid sharing};
\end{scope}

% --- Stage 4: Remask ---
\begin{scope}[shift={(10.61, \cy)}]
  \begin{scope}
    \clip[rounded corners=5pt] (0,0) rectangle (\cw,\ch);
    \fill[white] (0,0) rectangle (\cw,\ch);
    \fill[f2s4bd] (0, \ch-0.11) rectangle (\cw, \ch);
  \end{scope}
  \draw[rounded corners=5pt, line width=0.75pt, color=f2cardbd] (0,0) rectangle (\cw,\ch);
  \node[font=\rmfamily\bfseries\small, text=f2ink] at (\cw/2, 2.43) {Remask};
  \pic at (0.94, 2.00) {f2watch=f2s4bd};
  \node[rounded corners=3pt, draw=f2s4bd!65, fill=f2s4pill, line width=0.55pt,
        font=\rmfamily\bfseries\scriptsize, text=f2s4bd, inner xsep=5.5pt, inner ysep=2.2pt]
        at (1.78, 2.00) {78 ms};
  % Icon: 4x4 grid with neutral existing cells + stage-highlighted repaired cell
  \begin{scope}[shift={(1.12, 0.86)}]
    \fill[white] (0,0) rectangle (0.80,0.80);
    \fill[f2gridmute] (0.20,0.60) rectangle (0.40,0.80);
    \fill[f2gridmute] (0.60,0.40) rectangle (0.80,0.60);
    \fill[f2s4bd!24] (0.40,0.20) rectangle (0.60,0.40);
    \draw[line width=0.55pt, color=f2nodebd] (0,0) grid[step=0.20] (0.80,0.80);
    \draw[line width=0.95pt, color=f2s4bd] (0.40,0.20) rectangle (0.60,0.40);
    \draw[->, line width=0.8pt, color=f2s4bd] (0.96,0.52) arc[start angle=30,end angle=-55,radius=0.25];
  \end{scope}
  \node[align=center, font=\rmfamily\scriptsize, text=f2muted, inner sep=1pt]
    at (\cw/2, 0.36) {Remaining-demand masks\\and one-resource repair};
\end{scope}

% --- Stage 5: Pair ---
\begin{scope}[shift={(14.04, \cy)}]
  \begin{scope}
    \clip[rounded corners=5pt] (0,0) rectangle (\cw,\ch);
    \fill[white] (0,0) rectangle (\cw,\ch);
    \fill[f2s5bd] (0, \ch-0.11) rectangle (\cw, \ch);
  \end{scope}
  \draw[rounded corners=5pt, line width=0.75pt, color=f2cardbd] (0,0) rectangle (\cw,\ch);
  \node[font=\rmfamily\bfseries\small, text=f2ink] at (\cw/2, 2.43) {Pair};
  \pic at (0.94, 2.00) {f2watch=f2s5bd};
  \node[rounded corners=3pt, draw=f2s5bd!65, fill=f2s5pill, line width=0.55pt,
        font=\rmfamily\bfseries\scriptsize, text=f2s5bd, inner xsep=5.5pt, inner ysep=2.2pt]
        at (1.78, 2.00) {84 ms};
  % Icon: two 3x3 grids with bidirectional ruin-and-recreate arrows
  \begin{scope}[shift={(0.55, 0.92)}]
    \fill[white] (0,0) rectangle (0.60,0.60);
    \fill[f2gridmute] (0.20,0.20) rectangle (0.40,0.40);
    \draw[line width=0.55pt, color=f2nodebd] (0,0) grid[step=0.20] (0.60,0.60);
    \fill[white] (1.34,0) rectangle (1.94,0.60);
    \fill[f2s5bd!22] (1.54,0.20) rectangle (1.74,0.40);
    \draw[line width=0.55pt, color=f2nodebd] (1.34,0) grid[step=0.20] (1.94,0.60);
    \draw[line width=0.85pt, color=f2s5bd] (1.54,0.20) rectangle (1.74,0.40);
    \draw[->, line width=0.9pt, color=f2s5bd] (0.68,0.45) to[bend left=28] (1.26,0.45);
    \draw[->, line width=0.9pt, color=f2s5bd] (1.26,0.15) to[bend left=28] (0.68,0.15);
  \end{scope}
  \node[align=center, font=\rmfamily\scriptsize, text=f2muted, inner sep=1pt]
    at (\cw/2, 0.36) {Two-resource\\ruin-and-recreate};
\end{scope}

% Arrows connecting the 5 stages
\foreach \xstart in {3.40, 6.83, 10.26, 13.69} {
  \draw[f2stagearrow] (\xstart, \cy + \ch/2) -- (\xstart + 0.31, \cy + \ch/2);
}

% Bottom budget summary strip inside Panel (a)
\draw[rounded corners=3pt, line width=0.55pt, color=f2canvasbd, fill=white]
  (0.32, 0.22) rectangle (17.08, 0.68);
\node[font=\rmfamily\scriptsize, text=f2muted] at (8.70, 0.45)
  {\textcolor{f2ink}{\textbf{\(\Gamma = 87\text{ ms}\)}}\enskip\textcolor{f2canvasbd}{\(\vert\)}\enskip
   Search cutoff: 84 ms; tail: 3 ms\enskip\textcolor{f2canvasbd}{\(\vert\)}\enskip
   Includes input parsing\enskip\textcolor{f2canvasbd}{\(\vert\)}\enskip
   Cumulative cutoffs\enskip\textcolor{f2canvasbd}{\(\vert\)}\enskip
   \textcolor{f2ink}{\textbf{Full}} = Base + Global + CG + Remask + Pair};

% =========================================================================
% PANEL (b): Shared commit-or-discard gate (width = 17.4 cm, height = 6.90 cm)
% =========================================================================
\begin{scope}[shift={(0, -7.22)}]
\draw[rounded corners=6pt, dashed, line width=0.8pt, color=f2canvasbd]
  (0, 0) rectangle (17.4, 6.90);

\node[anchor=west, font=\rmfamily\bfseries\small, text=f2ink]
  at (0.32, 6.54) {(b) Shared commit-or-discard gate};

% Launch-policy pill (neutral cool slate -- avoids extra competing hue)
\node[anchor=east, rounded corners=4pt, draw=f2canvasbd, fill=f2badgebg, line width=0.65pt,
      font=\rmfamily\scriptsize, text=f2muted, inner xsep=9pt, inner ysep=3pt]
  (policy) at (17.08, 6.54)
  {Launch stage \textcolor{f2ink}{\(j\)} only if prerequisites hold and \textcolor{f2ink}{\(\tau_{\mathrm{in}} < t_j\)}.};

% --- Main refinement loop nodes (centered around y = 4.38) ---
% 1. Initialize I <- I_0 (clean neutral white node)
\node[rounded corners=4pt, draw=f2nodebd, fill=white, line width=0.8pt,
      minimum width=2.48cm, minimum height=1.26cm, align=center,
      font=\rmfamily\scriptsize, text=f2ink]
  (init) at (1.58, 4.38)
  {\textbf{Initialize \(I \leftarrow I_0\)}\\[3pt]
   \textcolor{f2muted}{Empty, complete,}\\
   \textcolor{f2muted}{feasible fallback}};

% 2. Stage / candidate controller diamond (clean neutral white node)
\node[diamond, aspect=1.32, draw=f2nodebd, fill=white, line width=0.85pt,
      inner sep=2pt, align=center, font=\rmfamily\scriptsize, text=f2ink]
  (ctrl) at (4.98, 4.38)
  {\textbf{Stage /}\\[-1pt]
   \textbf{controller}\\[2pt]
   \textcolor{f2muted}{Eligible work}\\[-1pt]
   \textcolor{f2muted}{remaining?}};

% 3. Build isolated candidate C (isolation expressed via dashed neutral frame)
\node[rounded corners=4pt, dashed, draw=f2nodebd, fill=white, line width=0.9pt,
      minimum width=2.66cm, minimum height=1.82cm, align=center,
      font=\rmfamily\scriptsize, text=f2ink]
  (build) at (8.55, 4.38)
  {\textbf{Build isolated}\\[-1pt]
   \textbf{candidate \(C\)}\\[26pt]
   \textcolor{f2muted}{\tiny \(I\) stays unchanged}\\[-1pt]
   \textcolor{f2muted}{\tiny during search}};

% Mini grid icon inside Build box (restrained cool slate)
\begin{scope}[shift={(8.25, 4.06)}]
  \fill[white] (0,0) rectangle (0.60,0.60);
  \fill[f2gridmute] (0,0.40) rectangle (0.20,0.60);
  \fill[f2gatebd!22] (0.20,0.20) rectangle (0.40,0.40);
  \fill[f2gridmute] (0.40,0) rectangle (0.60,0.20);
  \draw[line width=0.55pt, color=f2nodebd] (0,0) grid[step=0.20] (0.60,0.60);
\end{scope}

% 4. Common transactional gate (PRIMARY VISUAL FOCUS: deep academic blue)
\node[rounded corners=5pt, draw=f2gatebd, fill=f2gatebg, line width=1.1pt,
      minimum width=3.68cm, minimum height=2.04cm, align=center,
      font=\rmfamily\scriptsize, text=f2ink, inner xsep=7pt, inner ysep=5pt]
  (gate) at (12.22, 4.38)
  {\textcolor{f2gatebd}{\textbf{Common transactional gate}}\\[4pt]
   \begin{tabular}{@{}l@{\ \ }l@{}}
     Complete: & \textcolor{f2gatebd}{\(\operatorname{complete}(C)\)}\\[1pt]
     Feasible: & \textcolor{f2gatebd}{\(V(C) = 1\)}\\[1pt]
     Timely: & \textcolor{f2gatebd}{\(\tau_{\mathrm{done}}(C) < t_j\)}\\[1pt]
     Improving: & \textcolor{f2gatebd}{\(F(C) > F(I)\)}
   \end{tabular}\\[4pt]
   \textcolor{f2muted}{\tiny \(\tau_{\mathrm{done}}\) is measured after validation.}};

% 5. PASS ALL & FAIL ANY (exact semantic pair from manuscript tables)
\node[rounded corners=4pt, draw=f2passbd, fill=f2passbg, line width=0.85pt,
      minimum width=2.36cm, minimum height=0.82cm, align=center,
      font=\rmfamily\scriptsize, text=f2passtx]
  (pass) at (15.72, 4.92)
  {\textbf{\(\checkmark\)\ PASS ALL}\\[1pt]
   Commit: \(I \leftarrow C\)};

\node[rounded corners=4pt, draw=f2failbd, fill=f2failbg, line width=0.85pt,
      minimum width=2.36cm, minimum height=0.82cm, align=center,
      font=\rmfamily\scriptsize, text=f2failtx]
  (fail) at (15.72, 3.82)
  {\textbf{\(\times\)\ FAIL ANY}\\[1pt]
   Discard \(C\); keep \(I\)};

% Arrows in Row 1
\draw[f2arrow] (init.east) -- (ctrl.west);
\draw[f2arrow] (ctrl.east) -- node[above, font=\rmfamily\bfseries\scriptsize, text=f2ink] {Yes} (build.west);
\draw[f2arrow] (build.east) -- (gate.west);

% Split from gate to PASS / FAIL
\coordinate (split) at ($(gate.east)+(0.24,0)$);
\draw[line width=0.9pt, color=f2ink] (gate.east) -- (split);
\draw[f2arrow] (split) |- (pass.west);
\draw[f2arrow] (split) |- (fail.west);

% Feedback loop back to controller (y = 5.84, clear of top pills)
\draw[line width=0.85pt, color=f2nodebd] (pass.east) -- (17.16, 4.92);
\draw[line width=0.85pt, color=f2nodebd] (fail.east) -- (17.16, 3.82) -- (17.16, 5.84);
\draw[f2stagearrow] (17.16, 5.84) -- node[above, pos=0.66, font=\rmfamily\scriptsize, text=f2muted] {Next attempt / stage}
             (4.98, 5.84) -- (ctrl.north);
\node[font=\rmfamily\scriptsize, text=f2muted] at (3.20, 5.84) {Skip expired stages};

% --- Reserved 3-ms tail band (calm cool-slate strip) ---
\draw[rounded corners=5pt, dashed, line width=0.8pt, color=f2tailbd, fill=f2tailbg]
  (0.32, 0.84) rectangle (17.08, 2.74);

\node[anchor=west, align=left, font=\rmfamily\bfseries\scriptsize, text=f2ink]
  at (0.56, 2.12) {Final validation\\\& serialization};
\node[anchor=west, rounded corners=3pt, draw=f2tailbd, fill=white, line width=0.55pt,
      font=\rmfamily\bfseries\scriptsize, text=f2muted, inner xsep=5pt, inner ysep=2.2pt]
  at (0.56, 1.32) {Reserved 3-ms tail};

% Final check diamond
\node[diamond, aspect=1.95, draw=f2nodebd, fill=white, line width=0.85pt,
      inner sep=2pt, align=center, font=\rmfamily\scriptsize, text=f2ink]
  (finalchk) at (4.98, 1.79) {\(\operatorname{complete}(I)\)\\[-1pt]\(\land\; V(I)=1\)?};

% Arrow from controller (No) cleanly down to finalchk
\draw[f2arrow] (ctrl.south) -- node[left, pos=0.45, font=\rmfamily\bfseries\scriptsize, text=f2ink] {No} (finalchk.north);

% Yes branch: Serialize and output I (primary output tied subtly to gate blue)
\node[rounded corners=4pt, draw=f2gatebd, fill=white, line width=0.85pt,
      minimum width=3.60cm, minimum height=0.60cm, align=left,
      font=\rmfamily\scriptsize, text=f2ink, inner xsep=6pt]
  (outI) at (10.35, 2.22) {\hspace{0.34cm}\textbf{Serialize and output \(I\)}};
\pic at (8.90, 2.22) {f2doc=f2gatebd};

% No branch: Restore I <- I_0 -> Serialize and output I_0 (neutral fallback)
\node[rounded corners=4pt, draw=f2nodebd, fill=white, line width=0.8pt,
      minimum width=2.65cm, minimum height=0.60cm, align=center,
      font=\rmfamily\scriptsize, text=f2ink]
  (restore) at (9.88, 1.36) {Restore \(I \leftarrow I_0\)};

\node[rounded corners=4pt, draw=f2nodebd, fill=white, line width=0.8pt,
      minimum width=3.68cm, minimum height=0.60cm, align=left,
      font=\rmfamily\scriptsize, text=f2ink, inner xsep=6pt]
  (outI0) at (14.65, 1.36) {\hspace{0.34cm}\textbf{Serialize and output \(I_0\)}};
\pic at (13.16, 1.36) {f2doc=f2nodebd};

% Arrows inside reserved tail
\coordinate (tailsplit) at ($(finalchk.east)+(0.55,0)$);
\draw[line width=0.9pt, color=f2ink] (finalchk.east) -- (tailsplit);
\draw[f2arrow] (tailsplit) |- node[above, pos=0.72, font=\rmfamily\bfseries\scriptsize, text=f2passtx] {Yes} (outI.west);
\draw[f2arrow] (tailsplit) |- node[below, pos=0.72, font=\rmfamily\bfseries\scriptsize, text=f2failtx] {No} (restore.west);
\draw[f2arrow] (restore.east) -- (outI0.west);

% --- Bottom callout pills inside Panel (b) (quiet, non-competing) ---
\node[anchor=west, rounded corners=4pt, draw=f2passbd!55, fill=f2passbg, line width=0.6pt,
      font=\rmfamily\scriptsize, text=f2ink, inner xsep=10pt, inner ysep=3.5pt]
  at (0.32, 0.39) {\textcolor{f2passtx}{\textbf{Invariant:}}\ A complete feasible incumbent is retained throughout refinement.};

\node[anchor=east, rounded corners=4pt, draw=f2canvasbd, fill=f2badgebg, line width=0.6pt,
      font=\rmfamily\scriptsize, text=f2muted, inner xsep=10pt, inner ysep=3.5pt]
  at (17.08, 0.39) {\textcolor{f2ink}{\textbf{100-ms deadline:}}\ external evaluation only.};

\end{scope}
\end{tikzpicture}%
}%
\endgroup
\caption{Deadline-aware FPTR pipeline. (a) Ordered stages operate under cumulative completion cutoffs within the 87-ms internal budget. (b) An isolated candidate replaces the incumbent only after timely completion, validation, and strict improvement; otherwise, the feasible incumbent is retained. Final validation and serialization run in the reserved 3-ms tail.}
\label{fig:pipeline}
\end{figure*}
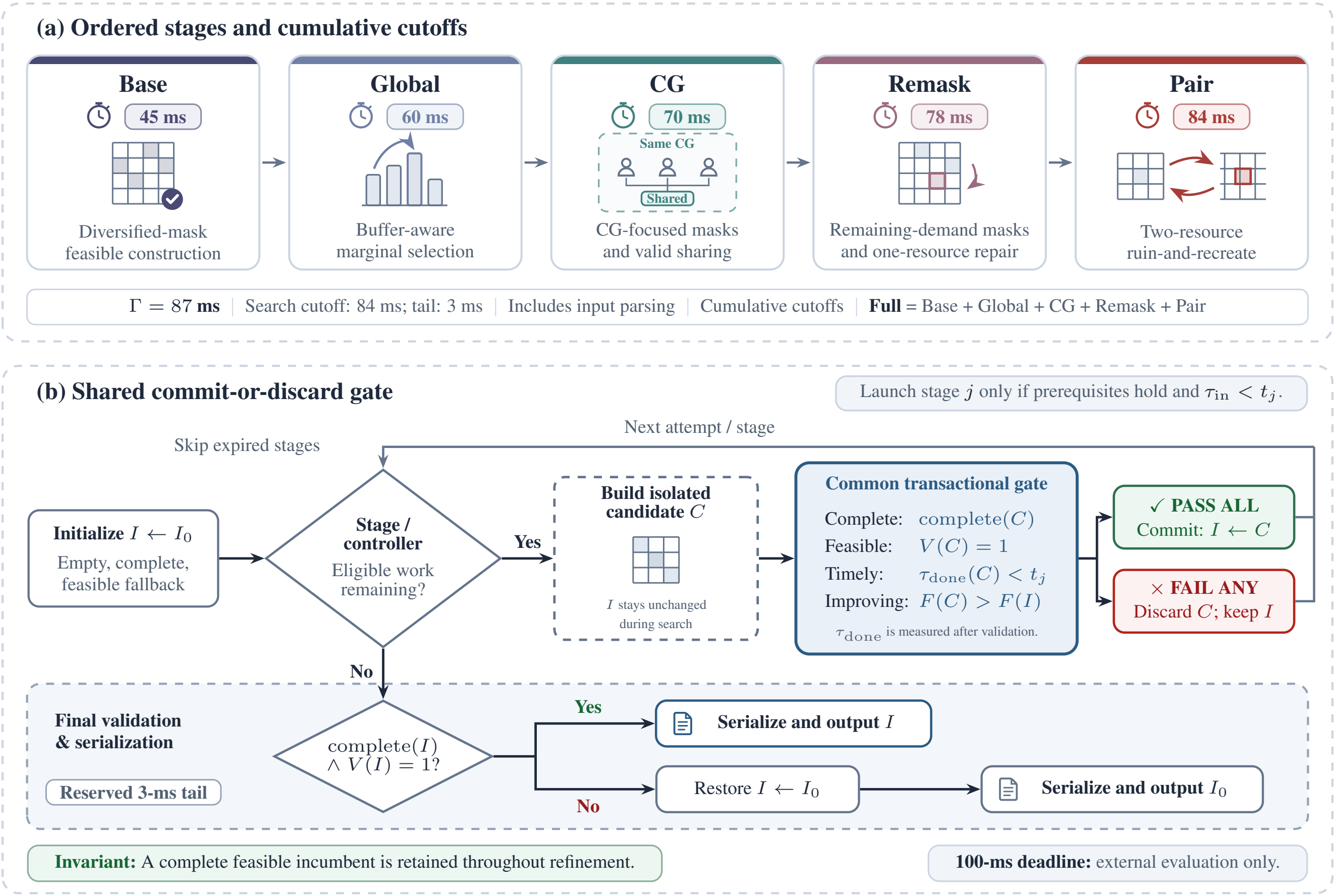

FPTR maintains two core states: an unconditional, complete, feasible fallback \(I_0\) (the empty allocation) and a releasable feasible incumbent \(I\) that can be output at any point.
Refinement proceeds through \(J=5\) ordered stages; each bounded operator \(\mathcal R_j\) operates under a cumulative stage cutoff \(t_j\) and constructs isolated candidates \(C\) without mutating \(I\).
Depending on the stage, \(\mathcal R_j\) either modifies a copy of \(I\) or builds \(C\) from stage-specific beam masks.
An allocation is \emph{complete} if every required decision field is populated, and \emph{feasible} if it satisfies all constraints in Section~\ref{sec:model}.
An internal validator \(V(A)\) returns \(1\) if allocation \(A\) satisfies every model constraint and \(0\) otherwise.
Let \(\tau_{\mathrm{done}}(C)\) denote the elapsed time recorded immediately after \(V(C)\) returns, when commit eligibility is decided; candidate \(C\) is \emph{timely} if \(\tau_{\mathrm{done}}(C)<t_j\).
Formally, FPTR is specified by the tuple \((I_0,\{\mathcal R_j,t_j\}_{j=1}^{J},V,F)\), where \(F\) is the clipped-traffic objective in~\eqref{eq:objective}.
Figure~\ref{fig:pipeline} illustrates the stage schedule and commit-or-discard flow.

\subsection{Algorithm and Cumulative Configurations}

Algorithm~\ref{alg:fptr} and Fig.~\ref{fig:pipeline}(b) depict the shared commit-or-discard control flow, while Fig.~\ref{fig:pipeline}(a) summarizes the cumulative cutoffs.
Execution begins with the empty fallback \(I\leftarrow I_0\).
Both Base and every subsequent refinement stage construct isolated candidates---a stage may propose multiple bounded candidates---and the incumbent is updated solely through the common gate~\eqref{eq:gate}.

\begin{algorithm}[t]
\caption{FPTR commit-or-discard refinement}
\label{alg:fptr}
\footnotesize
\setlength{\tabcolsep}{1.5pt}
\begin{tabular}{@{}r@{\ }>{\raggedright\arraybackslash}p{\dimexpr\columnwidth-1.8em\relax}@{}}
1: & Initialize \(I\leftarrow I_0\), where \(I_0\) is the empty complete feasible allocation.\\
2: & \textbf{for each stage} \(j\in\{\mathrm{Base},\mathrm{Global},\mathrm{CG},\mathrm{Remask},\mathrm{Pair}\}\) \textbf{do}\\
3: & \hangindent=1em\hangafter=0\hspace*{1em}Record stage-entry time \(\tau_{\mathrm{in}}\); \textbf{if} \(\tau_{\mathrm{in}}\ge t_j\) \textbf{then continue}.\\
4: & \hangindent=1em\hangafter=0\hspace*{1em}Construct isolated candidate \(C\leftarrow\mathcal R_j(I)\) without modifying \(I\). If \(C\) is complete, strictly improving, and before \(t_j\), evaluate \(V(C)\) and record \(\tau_{\mathrm{done}}(C)\) when it returns.\\
5: & \hangindent=1em\hangafter=0\hspace*{1em}\textbf{if} \(C\) passes~\eqref{eq:gate} \textbf{then} \(I\leftarrow C\); otherwise discard \(C\).\\
6: & \textbf{end for}\\
7: & \textbf{if} \(\neg\operatorname{complete}(I)\) or \(V(I)=0\) \textbf{then} restore \(I\leftarrow I_0\).\\
8: & Serialize and \textbf{return} \(I\).
\end{tabular}
\end{algorithm}

The stage-entry check decides whether a stage launches, whereas the post-validation timestamp check in the gate enforces timeliness.
If Base is skipped or rejected, \(I_0\) remains available; before serialization, a final validation check restores \(I_0\) if any anomaly occurs.

To isolate incremental contributions, we evaluate six configurations (Table~\ref{tab:configs}): BeamFirst is a lightweight standalone reference, whereas Base through Full form a cumulative ladder ending with Pair (the fifth and final stage).

\begin{table}[t]
\centering
\scriptsize
\renewcommand{\arraystretch}{1.12}
\setlength{\tabcolsep}{3.5pt}
\begin{tabular}{@{}l>{\raggedright\arraybackslash}p{4.65cm}r@{}}
\toprule
\rowcolor{tblhead}
\textbf{Config.} & \textbf{Cumulative stage added and purpose} & \textbf{\(t_j\) (ms)}\\
\midrule
\(\circ\)\ \textit{BeamFirst} & \textit{independent aggregate-mask reference} & \textcolor{tblmuted}{--}\\
\midrule
\(\triangleright\)\ Base & diversified-mask feasible construction & 45\\
\rowcolor{tblstripe}
\(+\)\ Global & \(+\) buffer-aware marginal selection & 60\\
\(+\)\ CG & \(+\) CG-focused masks and valid sharing & 70\\
\rowcolor{tblstripe}
\(+\)\ Remask & \(+\) demand masks and one-resource repair & \underline{78}\\
\rowcolor{tblhi}
\(\star\)\ \textbf{Full} & \textbf{\(+\) two-resource ruin-and-recreate (\textit{Pair})} & \textbf{84}\\
\bottomrule
\end{tabular}
\caption{Cumulative configurations and stage cutoffs for \(\Gamma=87\)~ms. \textit{BeamFirst} (\(\circ\)) is a standalone reference; Base (\(\triangleright\)) through \textbf{Full} form the cumulative FPTR ladder.}
\label{tab:configs}
\end{table}

The five cumulative stages target complementary bottlenecks.
\textbf{Base} generates a diverse pool of budget-feasible beam masks and sequentially assigns enabled resources using each user's remaining buffer.
\textbf{Global} revisits the top-ranked masks retained from Base and iteratively assigns whichever free resource--group pair yields the largest buffer-clipped gain.
\textbf{CG} constructs group-weighted beam masks and applies global buffer-aware allocation to unlock multi-user sharing.
\textbf{Remask} rebuilds beam masks from the incumbent's active users and residual demand, followed by bounded single-resource reassignment passes.
Finally, \textbf{Pair} performs bounded two-resource ruin-and-recreate, emptying pairs of enabled resources and refilling each pair in both orders to escape local optima.

\subsection{Exact Local CG Primitive}

Naively choosing the best multi-user subset \(G_r\subseteq\CG_m\) on a resource requires enumerating \(2^{|\CG_m|}\) subsets.
However, once the beam mask \(\alpha\), an unassigned or emptied resource \(r\), the accumulated traffic from all other resources, and the sharing cardinality \(s=|G_r|\) are fixed, each user's marginal gain becomes independent of which other users share the resource.
Formally, let \(Y_u=\sum_{\kappa\ne r:\,u\in G_\kappa}q(z_{u,\kappa})\) denote the unclipped traffic that user \(u\) accumulates from all resources other than \(r\).
For a fixed CG \(\CG_m\) and sharing cardinality \(s\), define the buffer-aware marginal gain
\begin{equation}
\begin{aligned}
  \delta_{u,r}^{(s)}
  ={}&\min\{B_u,Y_u+q(z_{u,r}^{(s)})\}\\
  &-\min\{B_u,Y_u\},
\end{aligned}
\label{eq:marginal}
\end{equation}
where \(z_{u,r}^{(s)}\) evaluates~\eqref{eq:effective-link} with \(|G_r|=s\).

\begin{proposition}[Top-\(s\) CG selection at fixed cardinality]
For fixed \(\alpha\), an unassigned or emptied resource \(r\), retained traffic \(Y\) on the other resources, group \(\CG_m\), and integer \(2\le s\le|\CG_m|\), a size-\(s\) subset \(S\subseteq\CG_m\) that maximizes incremental traffic is formed by selecting the \(s\) users with the largest \(\delta_{u,r}^{(s)}\).
\end{proposition}

\begin{proof}
With \(s\) fixed, the cardinality penalty \(10\log_{10}(1/s)\) is identical across all size-\(s\) subsets \(S\subseteq\CG_m\).
Because each user's buffer clipping depends only on its own \(Y_u\) and \(q(z_{u,r}^{(s)})\), the incremental traffic of \(S\) decouples as \(\sum_{u\in S}\delta_{u,r}^{(s)}\), which is maximized by the \(s\) largest terms.
\end{proof}

Let \(g_m=|\CG_m|\).
Including the singleton scan over \(\Users\) and scanning all cardinalities \(2\le s\le g_m\) per group, one fixed-resource, fixed-mask query runs in \(O(N+\sum_{m=1}^{M} g_m^2\log g_m)\) time---replacing exponential subset enumeration with a polynomial scan.
When Global scans all \(K_{\mathrm{free}}\) free enabled resources, this cost scales by a factor of \(K_{\mathrm{free}}\).

\subsection{Transactional Gating and Feasibility Invariant}

Let \(I\) be the current feasible incumbent and \(C\) an isolated candidate produced in stage \(j\).
Candidate \(C\) replaces \(I\) if and only if
\begin{equation}
\begin{aligned}
  &\operatorname{complete}(C)\land \tau_{\mathrm{done}}(C)<t_j\\
  &\quad\land V(C)=1\land F(C)>F(I),
\end{aligned}
\label{eq:gate}
\end{equation}
where \(\tau_{\mathrm{done}}(C)\) is the elapsed time measured immediately after \(V(C)\) returns.
The validator \(V\) checks index bounds, the global beam budget, explicit resource memberships, subband activation, per-resource user cardinality, and CG compatibility.
In practice, the conjunction may be short-circuited so that incomplete, expired, or non-improving candidates skip \(V\).
Crucially, neither entering stage \(j\) nor finishing candidate construction before \(t_j\) is sufficient on its own: validation itself must also complete before \(t_j\).
Together, the four predicates reject partial, late, infeasible, and non-improving candidates.
As noted in Section~\ref{sec:intro}, ``transactional'' denotes only this whole-candidate commit-or-discard rule.

\begin{lemma}[Feasible-incumbent invariant]
Assume that \(I_0\) is feasible and that validator \(V\) is sound for the constraints in Section~\ref{sec:model}.
Then incumbent \(I\) is feasible after every gate operation and immediately prior to serialization.
\end{lemma}

\begin{proof}
The base case holds because \(I\) is initialized to the feasible fallback \(I_0\).
At each gate evaluation, rejection leaves \(I\) unchanged, whereas acceptance requires \(V(C)=1\); by induction, \(I\) remains feasible after every gate operation.
Finally, the pre-serialization check retains \(I\) only if it is complete and passes \(V(I)=1\), restoring \(I_0\) otherwise.
\end{proof}

\subsection{Runtime Budgeting and Termination Behavior}

Let \(\Gamma\) denote the total internal time budget.
The solver starts its clock before input parsing and reserves a tail margin of \(\tau_{\mathrm{tail}}=3\)~ms for final validation and serialization whenever \(\Gamma\ge20\)~ms.
Using the reference cumulative cutoffs \(\rho=(45,60,70,78,84)\)~ms from Table~\ref{tab:configs}, stage \(j\) enforces the scaled cutoff \(t_j=(\Gamma-\tau_{\mathrm{tail}})\rho_j/84\), which adapts proportionally to the search window \(\Gamma-\tau_{\mathrm{tail}}\) and recovers \(\rho_j\) identically at \(\Gamma=87\)~ms.
Stage-entry checks prevent expired stages from launching, while the post-validation check in~\eqref{eq:gate} blocks late candidates from committing.

\section{Experimental Evaluation}
\label{sec:eval}

We organize our evaluation around five questions: internal cumulative traffic gains (Q1), matched external-baseline comparisons (Q1b), observed feasibility and runtime (Q2), quality--time trade-offs and CG stress (Q3), and optimality gaps on small exact cases (Q4).

\subsection{Protocol}

All single-threaded C++17 binaries (\texttt{g++} 13.3, \texttt{-O2}) ran on Linux 6.8 with a 2.30-GHz Intel Xeon E5-2686 v4 CPU.
External wall time was measured from subprocess launch through output collection; any nonzero exit code, infeasible output, or run exceeding the 100-ms deadline received zero credited traffic.
An independent Python validator reparsed every emitted allocation, checked all model constraints, and recomputed the objective in~\eqref{eq:objective}.

Table~\ref{tab:scenarios} summarizes the five fixed synthetic regimes, each comprising 30 consecutive seeds (150 instances in total).
Across regimes, the target fraction of users placed in CGs spans 0.35 to 0.75, and in the last regime 80\% of dual-membership resources have nonadjacent subbands.
Every instance--configuration pair was executed across five independent, randomized process launches.
Let \(\widetilde F_{i,m}=\operatorname{median}_{j=1}^{5}F_{i,m,j}\) be the instance-level median credited traffic of method \(m\) on instance \(i\), and define the pairwise relative gain \(g_i(A,B)=(\widetilde F_{i,A}-\widetilde F_{i,B})/\widetilde F_{i,B}\) whenever \(\widetilde F_{i,B}>0\).
Reported percentages first average \(g_i\) within each regime and then weight all five regimes equally; hence they need not equal ratios of pooled mean traffic.
Confidence intervals (95\% CI) are computed via 5,000 scenario-stratified bootstrap samples using the same equal regime weights.
Win/tie/loss (W/T/L) counts report the number of instances (out of 150) on which Full's median traffic exceeds, equals, or trails the baseline's median.

The internal evaluation comprises 8,610 process launches across four suites (main comparison, budget sweep, CG stress, and exact calibration), where BeamFirst serves as a lightweight standalone reference and Base through Full measure cumulative refinement under one model.
To benchmark against independent search strategies, we additionally ran a matched external-baseline main suite on the identical 150 instances with five repeats, \(\Gamma=87\)~ms, the same validator, and the same 100-ms deadline (6,750 launches: nine methods---BeamFirst, Base, Full, and six external baselines---\(\times\) 150 instances \(\times\) five repeats).
The external baselines are independent C++17 implementations of adaptive large-neighborhood search (ALNS)~\cite{ropke2006}, tabu search~\cite{glover1989}, a genetic algorithm (GA)~\cite{srinivas1994}, simulated annealing (SA)~\cite{kirkpatrick1983}, iterated local search (ILS)~\cite{lorenco2003}, and GRASP~\cite{feo1995}, rather than reproductions of published domain-specific solvers.

\begin{table}[t]
\centering
\scriptsize
\renewcommand{\arraystretch}{1.10}
\setlength{\tabcolsep}{3.4pt}
\begin{tabular}{@{}lrrrrrr@{}}
\toprule
\rowcolor{tblhead}
\textbf{Regime} & \textbf{\(N\)} & \textbf{\(K\)} & \textbf{\(P\)} & \textbf{\(T\)} & \textbf{\(L\)} & \textbf{\(L/(PT)\)}\\
\midrule
\textit{Small balanced} & 20 & 18 & 8 & 4 & 14 & 0.44\\
\rowcolor{tblstripe}
\textit{Medium long-tail} & 50 & 36 & 16 & 9 & 46 & 0.32\\
\textit{Medium tight} & 50 & 36 & 16 & 9 & 26 & \underline{0.18}\\
\rowcolor{tblstripe}
\textit{Large mixed} & \textbf{100} & \textbf{72} & \textbf{32} & \textbf{18} & \textbf{127} & 0.22\\
\textit{Large nonadjacent}\(^{\dagger}\) & \textbf{100} & \textbf{72} & \textbf{32} & \textbf{18} & 92 & \textbf{0.16}\\
\bottomrule
\multicolumn{7}{@{}l@{}}{\scriptsize\textcolor{tblmuted}{\(^{\dagger}\)80\% of dual-membership resources use nonadjacent subbands.}}\\
\end{tabular}
\caption{Fixed main regimes (30 consecutive seeds each). Tightest and second-tightest beam-budget ratios \(L/(PT)\) are in \textbf{bold} and \underline{underlined}.}
\label{tab:scenarios}
\end{table}

\begin{figure*}[t]
\centering
\includegraphics[width=0.92\textwidth]{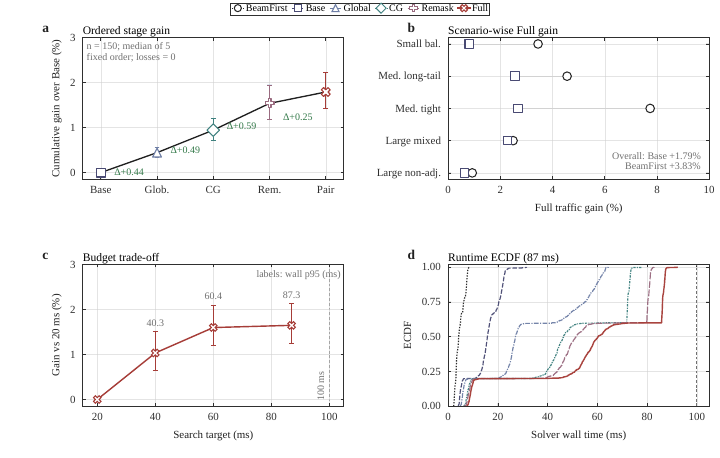}
\caption{Quality and runtime: (a) ordered stage gains, (b) regime-level Full gains, (c) budget trade-off, and (d) empirical CDF of solver wall time; error bars show 95\% stratified bootstrap intervals.}
\label{fig:quality-runtime}
\end{figure*}

\subsection{Main Results and Analysis}

\noindent\textbf{Q1: Does FPTR Yield Additional Feasible Traffic?}
At \(\Gamma=87\)~ms, Full increased mean instance-relative traffic by 3.83\% over BeamFirst (95\% CI 3.09--4.67\%) and by 1.79\% over Base (95\% CI 1.42--2.21\%), with W/T/L counts of 127/23/0 and 111/39/0, respectively.
Regime-level gains over Base were positive across all five regimes, ranging from 0.63\% to 2.68\% (Fig.~\ref{fig:quality-runtime}(b)).

Within the fixed stage order of Full, appending Global, CG, Remask, and Pair added 0.44\%, 0.49\%, 0.59\%, and 0.25\%, respectively (Fig.~\ref{fig:quality-runtime}(a)), strictly improving median traffic on 72, 34, 39, and 40 instances.
Because each stage starts from the prior incumbent, these values represent conditional increments along the pipeline rather than standalone module gains.

\medskip
\noindent\textbf{Q1b: Does Full Outperform Independent External Baselines?}
At the matched 87-ms budget, Full outperformed all six external metaheuristics by 6.68--8.35\% (Table~\ref{tab:external-comparison}).
Mean relative gains average over instances with nonzero baseline medians (excluding cases where repeated misses drove a baseline's median to zero), whereas W/T/L counts cover all 150 instances.
Full recorded zero 100-ms misses in 750 runs, versus 15--74 misses per external baseline (219 total).

\begin{table}[t]
\centering
\scriptsize
\renewcommand{\arraystretch}{1.10}
\setlength{\tabcolsep}{3.0pt}
\begin{tabular}{@{}lrrrr@{}}
\toprule
\rowcolor{tblhead}
\textbf{Method} & \textbf{W/T/L} & \textbf{Gain (\%)\,\(\uparrow\)} & \textbf{95\% CI (\%)} & \textbf{Misses\,\(\downarrow\)}\\
\midrule
\textit{ALNS} & 134/5/11 & \(+\)7.80 & 7.03--8.56 & \textcolor{tblwarn}{\textbf{74}}\\
\rowcolor{tblstripe}
\textit{Tabu} & \underline{139}/4/7 & \(+\)7.77 & 7.01--8.52 & \textcolor{tblwarn}{18}\\
\textit{GA} & 127/2/21 & \(+\)6.68 & 5.91--7.47 & \textcolor{tblwarn}{\underline{44}}\\
\rowcolor{tblstripe}
\textit{SA} & 137/4/9 & \underline{\(+\)7.90} & 7.19--8.58 & \textcolor{tblwarn}{15}\\
\textit{ILS} & 137/4/9 & \textbf{\(+\)8.35} & 7.53--9.13 & \textcolor{tblwarn}{41}\\
\rowcolor{tblstripe}
\textit{GRASP} & \textbf{140}/2/8 & \(+\)7.61 & 6.88--8.33 & \textcolor{tblwarn}{27}\\
\midrule
\rowcolor{tblhi}
\textbf{FPTR Full} & \textcolor{tblmuted}{\textit{ref.}} & \textcolor{tblmuted}{\textit{--}} & \textcolor{tblmuted}{\textit{--}} & \textcolor{tblgood}{\textbf{0} (\(\checkmark\))}\\
\bottomrule
\end{tabular}
\caption{Matched external-baseline comparison at \(\Gamma=87\)~ms (five-run medians; equal-weight regime-stratified gains; W/T/L of Full over 150 instances; misses out of 750 runs). Largest/second-largest margins, wins, and misses are in \textbf{bold}/\underline{underlined}.}
\label{tab:external-comparison}
\end{table}

\medskip
\noindent\textbf{Q2: What Feasibility and Runtime Were Observed?}
In the internal main suite, BeamFirst, Base, and Full achieved p50/p95 wall times of 4.23/7.71, 15.75/22.68, and 60.28/87.29~ms (Fig.~\ref{fig:quality-runtime}(d)).
All 750 Full launches passed validation and met the 100-ms deadline (worst wall time 92.14~ms).
Across all 8,610 internal launches, every output was valid with zero misses; in the external suite, all outputs were valid, but 219 missed 100~ms.

\begin{figure*}[t]
\centering
\includegraphics[width=0.88\textwidth]{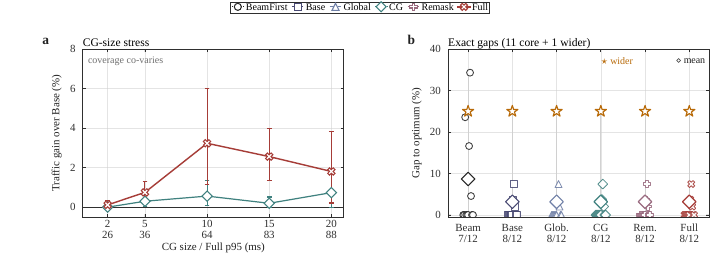}
\caption{Stress and calibration: (a) co-varying CG size and coverage, and (b) exact optimality gaps across 12 small cases.}
\label{fig:stress-optimality}
\end{figure*}

\medskip
\noindent\textbf{Q3: Where Do the Quality--Time Returns Diminish?}
Relative to 20~ms, Full gained 1.03\%, 1.60\%, and 1.65\% at \(\Gamma=40\), 60, and 87~ms (Fig.~\ref{fig:quality-runtime}(c)), adding only 0.05 percentage points past 60~ms.
Over the 20-to-87-ms sweep, the 87-ms gain has a 95\% CI of 1.24--2.14\% while p95 wall time rises from 20.44 to 87.31~ms.
On ten-instance CG stress sets at sizes \(2,5,10,15,20\), Full beat Base at every setting (Fig.~\ref{fig:stress-optimality}(a)), reaching 87.50/92.93~ms p95/worst wall time at size 20 with zero misses; because CG coverage co-varied with group size, this measures joint stress.

\medskip
\noindent\textbf{Q4: How Close Is FPTR to Exact Solutions on Small Cases?}
The exact calibration suite comprises eleven core cases with \((N,K,P,T)=(5,4,4,3)\) and one wider case with \((N,K,P,T)=(8,8,6,4)\), drawn from consecutive unfiltered seeds.
An independent dynamic-programming (DP) reference verified all 12 optima, and Base through Full produced identical objectives on every case (Fig.~\ref{fig:stress-optimality}(b)).
Full matched the exact optimum on 8 of the 12 cases, with a median optimality gap \((\mathrm{OPT}-F)/\mathrm{OPT}\) of 0\%, a mean gap of 3.19\%, and a maximum gap of 25\% (on the wider case).
By contrast, external ALNS, tabu search, and GA reached all 12 optima; SA, ILS, and GRASP ran only on the 87-ms main suite, so this calibration establishes no universal ranking.

\medskip
\noindent\textbf{Threats to Validity.}
Evaluation uses synthetic static snapshots, one unpinned host, and a small exact suite.
Although the external main suite matches inputs, objective, the 87-ms budget, validation, and 100-ms deadline accounting, its six metaheuristics are independent reference implementations rather than reproductions of published domain-specific solvers; conclusions remain bounded to this model, platform, and implementation family.
Moreover, the CG sweep co-varies group size with coverage, and the model omits dynamic multi-slot queues, fairness, mobility, detailed interference, continuous precoding, and deployment traces.
Observed runtimes are host-dependent rather than hard-real-time bounds, and the gate protects output only under normal termination---not crashes or forced preemption.

\section{Discussion and Conclusion}
\label{sec:conclusion}

FPTR delivers a consistent, though modest, traffic improvement over Base.
Because Base's diversified masks and buffer-aware greedy assignment already capture most broadly available gains, later stages target rarer residual structures under tighter remaining time.
This explains both selective stage acceptances and diminishing returns beyond 60~ms; nevertheless, strict-improvement gating prevents regression, letting Full beat Base on 111 of 150 cases with no losses.

The exact suite and external comparisons jointly clarify FPTR's guarantees.
Identical Base and Full scores mean no later candidate satisfied all four gate predicates---conservative gating, not failure---while the 25\% worst-case exact gap shows that FPTR guarantees feasible-incumbent preservation and monotonic improvement rather than near-optimality.
Reading the matched 87-ms main suite (Full leading all six external metaheuristics with zero deadline misses) alongside the 12-case exact suite (external ALNS, tabu search, and GA reaching all 12 optima versus 8 for Full) highlights the trade-off between deadline-safe refinement at scale and broader search on tiny cases; as independent implementations, these baselines do not generalize to all published solvers.

In conclusion, FPTR converts spare computation time into verified traffic gains while keeping a complete, validated allocation ready throughout execution.
Future work will address dynamic multi-slot queues, learned mask priors, and deployment traces under platform-specific timing budgets.

\medskip
\noindent\textbf{Acknowledgements.} This work was supported in part by the Major Key Project of Pengcheng Laboratory under Grant PCL2025A12-3, and in part by the Science and Technology Planning Project of Guangdong Science and Technology Department under the Guangdong Key Laboratory of Advanced IntelliSense Technology under Grant 2019B121203006.

\smallskip
\noindent\textbf{Code and data availability.} Source code: \href{https://github.com/rudykon/FPTR_Scheduler}{GitHub repository}; experimental data: \href{https://res-static.hc-cdn.cn/cloudbu-site/china/zh-cn/wuxian-gaoxiao2026/1780886490950118786.zip}{dataset archive}.

\phantomsection
\pdfbookmark[1]{References}{sec:references}
{
    \small
    \bibliographystyle{ieeenat_fullname}
    \bibliography{references}
}

\pdfbookmark[1]{Supplementary Material}{appendix:root}
\clearpage
\maketitlesupplementary
\appendix
\setcounter{section}{0}
\setcounter{table}{0}
\setcounter{equation}{0}
\renewcommand{\thetable}{A\arabic{table}}
\renewcommand{\theequation}{A\arabic{equation}}
\renewcommand{\theHsection}{appendix.\Alph{section}}
\renewcommand{\theHsubsection}{appendix.\Alph{section}.\arabic{subsection}}
\renewcommand{\theHtable}{appendix.\arabic{table}}
\renewcommand{\theHequation}{appendix.\arabic{equation}}

\section{Evidence and Reproducibility Summary}
\label{app:evidence}

This supplementary appendix summarizes execution accounting, run-level stability, stage acceptance, exact calibration, and artifact integrity. Unless stated otherwise, quality metrics use five-run instance medians, and execution counts denote independent process launches.

Rather than repeating the main results narrative, this companion record makes the statistical unit, crediting rule, and audit boundary explicit so that every reported figure can be traced directly to a process launch, a generated instance, or a checksum-sealed artifact.

\subsection{Reading Conventions and Statistical Units}

A \emph{launch} denotes a single independent subprocess invocation with its own input stream, randomized execution order, random seed (for stochastic baselines), and wall-clock measurement. An independent Python validator reparses every serialized allocation and recomputes the clipped transmitted-traffic objective; trace-reported scores are purely diagnostic and never substitute for independent recomputation.

For instance \(i\) and method \(m\), let \(F_{i,m,j}\) be the credited traffic from repeat \(j\). The summary statistics in the tables are defined over five-run instance medians:
\begin{equation}
\begin{aligned}
  \widetilde F_{i,m}&=\operatorname{median}_{j=1,\ldots,5}F_{i,m,j},\\
  g_i(A,B)&=\frac{\widetilde F_{i,A}-\widetilde F_{i,B}}{\widetilde F_{i,B}}.
\end{aligned}
\label{eq:app-median-gain}
\end{equation}
Pairwise relative gains \(g_i\) (defined when \(\widetilde F_{i,B}>0\)) are averaged within each regime before all five regimes receive equal weight, preventing the largest instances from dominating the overall summary by scale alone. A late yet structurally feasible output is preserved for diagnostics, but its credited traffic is zero under the 100-ms execution deadline.

\subsection{Execution Accounting and Configuration-Level Results}

\begin{table}[t]
\centering
\scriptsize
\renewcommand{\arraystretch}{1.10}
\setlength{\tabcolsep}{2.2pt}
\resizebox{\columnwidth}{!}{%
\begin{tabular}{@{}lrrrrrr@{}}
\toprule
\rowcolor{tblhead}
\textbf{Suite} & \textbf{Inst.} & \textbf{Cond.} & \textbf{Rep.} & \textbf{Launches} & \textbf{Valid\,\(\uparrow\)} & \textbf{Misses\,\(\downarrow\)}\\
\midrule
\textit{Main} & 150 & 6 & 5 & 4,500 & 4,500 & \textcolor{tblgood}{0}\\
\rowcolor{tblstripe}
\textit{Budget sweep} & 150 & 4 & 5 & 3,000 & 3,000 & \textcolor{tblgood}{0}\\
\textit{CG stress} & 50 & 3 & 5 & 750 & 750 & \textcolor{tblgood}{0}\\
\rowcolor{tblstripe}
\textit{Exact calib.} & 12 & 6 & 5 & 360 & 360 & \textcolor{tblgood}{0}\\
\midrule
\rowcolor{tblhi}
\textbf{Total} & \textcolor{tblmuted}{--} & \textcolor{tblmuted}{--} & \textcolor{tblmuted}{--} & \textbf{8,610} & \textbf{8,610} & \textcolor{tblgood}{\textbf{0} (\(\checkmark\))}\\
\bottomrule
\end{tabular}%
}
\caption{Formal internal launch accounting (Instances, Conditions, Repeats); all valid outputs passed independent verification (\(\checkmark\)).}
\label{tab:app-launches}
\end{table}

Table~\ref{tab:app-launches} summarizes all formal internal launches. Although the internal main and budget-sweep suites share the same 150 inputs, every condition was executed via fresh, independent process launches. The 500-ms test harness assigns zero credited traffic to any nonzero exit code, infeasible output, or wall time exceeding 100~ms.

Each internal suite answers a distinct question: the main suite evaluates cumulative configuration behavior; the budget sweep varies the internal search cap across the same workload family; the CG stress suite increases maximum group size and user coverage jointly; and the exact suite pairs heuristic runs with independent exact optimization on small instances. Their launch counts therefore represent complementary experimental dimensions rather than four disjoint input pools.

\subsection{Matched External-Baseline Suite}

The matched external main suite evaluates Full alongside six independent metaheuristics---ALNS, tabu search, GA, SA, ILS, and GRASP---on the identical 150 fixed inputs with five repeats, an 87-ms budget, the same validator, and the same 100-ms deadline (6,750 launches in total). Every output was structurally valid, whereas 219 external launches missed the 100-ms deadline. Earlier budget and exact calibration results for ALNS, tabu search, and GA remain archived under \texttt{reproducibility/external\_comparison\_extended/}.

Each external baseline is compiled as a standalone C++17 executable that shares only the input/output format, objective, and validation contract without calling FPTR's core scheduler or reusing its traces. Consequently, this comparison benchmarks independently implemented reference algorithms under a uniform accounting rule rather than claiming to reproduce all published solvers.

\begin{table*}[t]
\centering
\footnotesize
\renewcommand{\arraystretch}{1.12}
\setlength{\tabcolsep}{7pt}
\begin{tabular}{lrrrr}
\toprule
\rowcolor{tblhead}
\textbf{Method} & \textbf{W/T/L} & \textbf{Mean rel.\ gain (\%)\,\(\uparrow\)} & \textbf{95\% CI (\%)} & \textbf{Misses\,\(\downarrow\) (/750)}\\
\midrule
\textit{ALNS} & 134/5/11 & \(+\)7.80 & 7.03--8.56 & \textcolor{tblwarn}{\textbf{74}}\\
\rowcolor{tblstripe}
\textit{Tabu} & \underline{139}/4/7 & \(+\)7.77 & 7.01--8.52 & \textcolor{tblwarn}{18}\\
\textit{GA} & 127/2/21 & \(+\)6.68 & 5.91--7.47 & \textcolor{tblwarn}{\underline{44}}\\
\rowcolor{tblstripe}
\textit{SA} & 137/4/9 & \underline{\(+\)7.90} & 7.19--8.58 & \textcolor{tblwarn}{15}\\
\textit{ILS} & 137/4/9 & \textbf{\(+\)8.35} & 7.53--9.13 & \textcolor{tblwarn}{41}\\
\rowcolor{tblstripe}
\textit{GRASP} & \textbf{140}/2/8 & \(+\)7.61 & 6.88--8.33 & \textcolor{tblwarn}{27}\\
\midrule
\rowcolor{tblhi}
\textbf{FPTR Full (ours)} & \textcolor{tblmuted}{\textit{reference}} & \textcolor{tblmuted}{\textit{--}} & \textcolor{tblmuted}{\textit{--}} & \textcolor{tblgood}{\textbf{0} (\(\checkmark\))}\\
\bottomrule
\end{tabular}
\caption{Full versus six independent external baselines at the matched 87-ms main condition (\textbf{bold}/\underline{underline}: largest/second-largest margins, wins, and misses).}
\label{tab:app-external}
\end{table*}

As reported in Table~\ref{tab:app-external}, mean relative gains are averaged over instances with nonzero baseline medians: 141/150 pairs for ALNS, 149/150 for tabu search, 144/150 for GA, 150/150 for SA, 148/150 for ILS, and 149/150 for GRASP. W/T/L counts cover all 150 instance medians; the machine-readable archive resides at \texttt{reproducibility/external\_comparison\_v3/}.

\begin{table*}[t]
\centering
\footnotesize
\renewcommand{\arraystretch}{1.12}
\setlength{\tabcolsep}{6pt}
\begin{tabular}{lrrrrrr}
\toprule
\rowcolor{tblhead}
\textbf{Configuration} & \textbf{Mean \(\widetilde F\) \(\uparrow\)} & \textbf{Rel.\ Base (\%) \(\uparrow\)} & \textbf{p50 (ms) \(\downarrow\)} & \textbf{p95 (ms) \(\downarrow\)} & \textbf{Worst (ms) \(\downarrow\)} & \textbf{Misses \(\downarrow\)}\\
\midrule
\(\circ\)\ \textit{BeamFirst} & 13467.89 & \textcolor{tblwarn}{\(-\)1.82} & \textbf{4.23} & \textbf{7.71} & \textbf{8.82} & \textcolor{tblgood}{0/750} \\
\midrule
\(\triangleright\)\ Base & 13662.15 & \textcolor{tblmuted}{\(+\)0.00} & \underline{15.75} & \underline{22.68} & \underline{31.50} & \textcolor{tblgood}{0/750} \\
\rowcolor{tblstripe}
\(+\)\ Global & 13732.17 & \(+\)0.44 & 27.06 & 62.12 & 64.40 & \textcolor{tblgood}{0/750} \\
\(+\)\ CG & 13796.05 & \(+\)0.88 & 46.40 & 73.30 & 78.09 & \textcolor{tblgood}{0/750} \\
\rowcolor{tblstripe}
\(+\)\ Remask & \underline{13873.17} & \underline{\(+\)1.54} & 51.44 & 81.35 & 82.85 & \textcolor{tblgood}{0/750} \\
\rowcolor{tblhi}
\(\star\)\ \textbf{Full} & \textbf{13898.13} & \textbf{\(+\)1.79} & 60.28 & 87.29 & 92.14 & \textcolor{tblgood}{\textbf{0/750}} \\
\bottomrule
\end{tabular}
\caption{Main-suite quality and wall time across 150 five-run instance medians (\textbf{bold}: best; \underline{underline}: second-best per metric).}
\label{tab:app-config}
\end{table*}

Full achieved the highest mean instance-median traffic with p50/p95/worst wall times of 60.28/87.29/92.14~ms and zero misses in 750 launches (Table~\ref{tab:app-config}). Because Table~\ref{tab:app-config} weights regimes equally and pools 750 wall times without equalizing realized runtimes across stages, it characterizes quality--runtime operating points rather than compute-matched superiority.

Wall time spans subprocess launch, input transfer, solver execution, and output collection, while independent Python validation is timed separately. The 87-ms internal setting is a search budget rather than a hard completion-time guarantee; all latency percentiles reflect the tested host and implementation rather than a platform-independent upper bound.

\subsection{Run-Level Stability and Stage Acceptance}

Pairing each Full run with the Base or BeamFirst run sharing the exact same scenario, seed, and repeat index yields 750 same-repeat pairs. Against Base, Full won 564 pairs, tied 186, and lost none (unweighted mean relative gain: 1.800\%); against BeamFirst, it won 643, tied 107, and lost none (unweighted mean: 3.839\%). These run-level averages use a different unit and weighting than the instance-median, regime-weighted percentages reported in the main paper (Section~\ref{sec:eval}, Q1: 1.79\% and 3.83\%). Across repeats, all five Full launches yielded identical traffic on 122 of 150 instances; the relative five-run range (maximum minus minimum traffic divided by the instance median) had a 95th percentile of 4.969\% and a maximum of 7.592\%.

Repeated-process stability and stage-acceptance traces probe distinct properties. Repeated processes quantify run-to-run sensitivity to ordinary OS scheduling noise, whereas the acceptance trace records the fixed order Base--Global--CG--Remask--Pair inside each Full run, making every acceptance count conditional on the incumbent handed down by preceding stages.

\begin{table}[b]
\centering
\scriptsize
\renewcommand{\arraystretch}{1.10}
\setlength{\tabcolsep}{2.4pt}
\resizebox{\columnwidth}{!}{%
\begin{tabular}{@{}lrrrrr@{}}
\toprule
\rowcolor{tblhead}
\textbf{Scenario} & \textbf{\(n/5n\)} & \textbf{Global} & \textbf{CG} & \textbf{Remask} & \textbf{Pair}\\
\midrule
\textit{small-balanced} & 30/150 & 2/10 & 8/40 & 5/25 & 3/15 \\
\rowcolor{tblstripe}
\textit{medium-longtail} & 30/150 & \underline{18/90} & \textbf{11/55} & \textbf{17/85} & \textbf{23/115} \\
\textit{medium-tight} & 30/150 & 7/35 & 3/15 & \textbf{17/85} & \underline{14/70} \\
\rowcolor{tblstripe}
\textit{large-mixed} & 30/150 & \textbf{27/135} & \underline{10/49} & \textcolor{tblmuted}{\textit{0/0}} & \textcolor{tblmuted}{\textit{0/0}} \\
\textit{large-nonadjacent} & 30/150 & \underline{18/90} & 2/16 & \textcolor{tblmuted}{\textit{0/0}} & \textcolor{tblmuted}{\textit{0/0}} \\
\midrule
\rowcolor{tblhi}
\textbf{Overall} & \textbf{150/750} & \textbf{72/360} & 34/175 & 39/195 & \underline{40/200} \\
\bottomrule
\end{tabular}%
}
\caption{Fixed-order stage acceptance by scenario (instances/runs; \textbf{bold}/\underline{underline}: highest/second-highest per stage; zero counts muted).}
\label{tab:app-acceptance}
\end{table}

As detailed in Table~\ref{tab:app-acceptance}, overall instance/run acceptance rates were 48.0/48.0\% for Global, 22.7/23.3\% for CG, 26.0/26.0\% for Remask, and 26.7/26.7\% for Pair. Entries of 0/0 indicate that no candidate strictly improved the incumbent because of ties, stage expiry, or failed validation; all counts represent fixed-order conditional increments.

These rates should not be read as standalone module success probabilities. A downstream stage may record few acceptances because earlier stages have already captured the relevant residual structure, because little budget remains, or because candidates fail the shared gate. Even sparse acceptances reliably raise cumulative Full performance whenever every accepted update is feasible and strictly improving.

\subsection{Exact Calibration and Independent Audit}

\begin{table*}[t]
\centering
\footnotesize
\renewcommand{\arraystretch}{1.12}
\setlength{\tabcolsep}{5.2pt}
\begin{tabular}{lrrrrrclrrrrr}
\toprule
\rowcolor{tblhead}
\textbf{Case} & \textbf{\(F^\star\)} & \textbf{BF} & \textbf{Base} & \textbf{Full} & \textbf{Gap\(\downarrow\)} & & \textbf{Case} & \textbf{\(F^\star\)} & \textbf{BF} & \textbf{Base} & \textbf{Full} & \textbf{Gap\(\downarrow\)}\\
\midrule
\textbf{W01}\(^{\star}\) & 920 & 690 & 690 & 690 & \textcolor{tblwarn}{\textbf{25.00}} & & \textit{C07} & 449 & 449 & 449 & 449 & \textcolor{tblgood}{0.00}\\
\rowcolor{tblstripe}
\textit{C02} & 237 & 237 & 237 & 237 & \textcolor{tblgood}{0.00} & & \textit{C08} & 806 & 672 & 776 & 776 & \textcolor{tblwarn}{3.72}\\
\textit{C03} & 548 & 548 & 548 & 548 & \textcolor{tblgood}{0.00} & & \textit{C09} & 804 & 528 & 744 & 744 & \textcolor{tblwarn}{\underline{7.46}}\\
\rowcolor{tblstripe}
\textit{C04} & 1188 & 908 & 1188 & 1188 & \textcolor{tblgood}{0.00} & & \textit{C10} & 1665 & 1589 & 1631 & 1631 & \textcolor{tblwarn}{2.04}\\
\textit{C05} & 564 & 564 & 564 & 564 & \textcolor{tblgood}{0.00} & & \textit{C11} & 888 & 888 & 888 & 888 & \textcolor{tblgood}{0.00}\\
\rowcolor{tblstripe}
\textit{C06} & 648 & 648 & 648 & 648 & \textcolor{tblgood}{0.00} & & \textit{C12} & 888 & 888 & 888 & 888 & \textcolor{tblgood}{0.00}\\
\bottomrule
\end{tabular}
\caption{Exact calibration (BF = \textit{BeamFirst}; W01\(^{\star}\) is the wider case; seeds 20281001--20281012). Exact optima (\textcolor{tblgood}{0.00}) are green; largest and second-largest gaps (\%) are in \textbf{bold} and \underline{underlined}.}
\label{tab:app-exact}
\end{table*}

An independent dynamic-programming (DP) audit reproduced all 12 sealed optima (Table~\ref{tab:app-exact}) without invoking the formal solver: 390,814 attempted state transitions collapsed via 58,229 duplicates into 332,585 pre-prune states, of which dominance pruning removed 219,088 and retained 113,497. All layer-level and aggregate conservation identities held; this suite calibrates small-case quality only.

For each enumerated beam-mask plan, the audit DP maintains a vector of per-user accumulated traffic across resource layers, enumerates the empty choice and every legal compatibility-group subset, clips each vector at user buffers, and eliminates componentwise dominated states. Because this construction is completely independent of the C++ heuristic, the audited optimum serves as an objective calibration reference rather than a second heuristic baseline.

\newpage
\subsection{Budget, CG Stress, and Artifact Integrity}

Across the independent budget sweep at 20, 40, 60, and 87~ms, mean traffic gains relative to 20~ms were 0.00\%, 1.03\%, 1.60\%, and 1.65\%, while p95 wall times rose from 20.44 to 40.34, 60.39, and 87.31~ms; all 750 launches at each budget met the 100-ms deadline (with the 87-ms sweep run independently of the main suite). In the CG stress sweep across maximum group sizes 2, 5, 10, 15, and 20, CG user coverage and Full-over-Base relative gain co-varied from 6.7\% (gain +0.11\%) up to 66.7\% (gain +1.81\%), passing through intermediate gains of +0.75\%, +3.23\%, and +2.56\%; at size 20, Full recorded 87.50/92.93~ms p95/worst wall time and 0/50 deadline misses. SHA-256 manifests seal 14 formal result files, 16 exact-audit files (including 12 reconstructed inputs), and 21 derived appendix files. Note that hash-reconstructed inputs establish independence only for the exact optimizer's path, and the archive label \texttt{causal cumulative ablation} is retained verbatim for checksum compatibility to denote the fixed-order within-run trace.

The provenance chain is layered: case hashes bind generated inputs, source hashes bind generator and validator versions, and directory manifests seal all derived outputs. Any regenerated archive can therefore be audited for both numerical consistency and build provenance. Precompiled binaries are not treated as evidence; the documented build commands and compiler flags define the reproducible boundary.

\end{document}